\documentclass{article}
\usepackage{ijcai24}

\usepackage{times}
\usepackage{soul}
\usepackage{url}
\usepackage[hidelinks]{hyperref}
\usepackage[utf8]{inputenc}
\usepackage[small]{caption}
\usepackage{graphicx}
\usepackage{amsmath}
\usepackage{amsthm}
\usepackage{booktabs}
\usepackage{algorithm}
\usepackage[switch]{lineno}
\usepackage{booktabs}
\usepackage{multirow}
\usepackage{amsmath} 
\usepackage{algpseudocode}  
\usepackage{xcolor}
\title{LitE-SNN: Designing Lightweight and Efficient Spiking Neural Network through Spatial-Temporal Compressive Network Search and Joint Optimization}

\author{
Qianhui Liu$^1$
\and
Jiaqi Yan$^2$\and
Malu Zhang$^{3}$\and
Gang Pan$^{2*}$\And
Haizhou Li$^{4,5,1}$\thanks{Corresponding author}\\
\affiliations
$^1$National Univerisity of Singapore, Singapore \\
$^2$Zhejiang University, China\\
$^3$University of Electronic Science and Technology of China, China\\
$^4$The Chinese University of Hong Kong, Shenzhen (CUHK-Shenzhen), China\\
$^5$Shenzhen Research Institute of Big Data, China\\
\emails
qhliu@nus.edu.sg,
jiaqi\_yan@zju.edu.cn,
maluzhang@uestc.edu.cn,
gpan@zju.edu.cn,
haizhouli@cuhk.edu.cn
}
\begin{document}

\maketitle

\begin{abstract}	
Spiking Neural Networks (SNNs) mimic the information-processing mechanisms of the human brain and are highly energy-efficient, making them well-suited for low-power edge devices. However, the pursuit of accuracy in current studies leads to large, long-timestep SNNs, conflicting with the resource constraints of these devices. In order to design lightweight and efficient SNNs, we propose a new approach named LitE-SNN that incorporates both spatial and temporal compression into the automated network design process. Spatially, we present a novel Compressive Convolution block (CompConv) to expand the search space to support pruning and mixed-precision quantization. Temporally, we are the first to propose a compressive timestep search to identify the optimal number of timesteps under specific computation cost constraints. Finally, we formulate a joint optimization to simultaneously learn the architecture parameters and spatial-temporal compression strategies to achieve high performance while minimizing memory and computation costs. Experimental results on CIFAR-10, CIFAR-100, and Google Speech Command datasets demonstrate our proposed LitE-SNNs can achieve competitive or even higher accuracy with remarkably smaller model sizes and fewer computation costs. %Furthermore, we validate the effectiveness of our LitE-SNN on the trade-off between accuracy and resource cost and show the superiority of our joint optimization. Finally, we conduct energy analysis to further confirm the energy efficiency of LitE-SNN.
\end{abstract}

\section{Introduction}
Spiking Neural Networks (SNNs) have gained great attention in recent years \cite{liu2020unsupervised,zhang2021rectified,hu2021spiking,hu2023fast,wang2023event,wei2024event} due to their ability to mimic the information-processing mechanisms of the human brain.  They use the timing of the signals (spikes) to communicate between neuronal units. A unit in SNNs is only active when it receives or emits a spike, enabling event-driven processing and high energy efficiency. The received spikes are weight-accumulated into the membrane potential using only accumulate (AC) operations, which consume much lower energy than the standard energy-intensive multiply-accumulate (MAC) strategy in Artificial Neural Networks (ANNs) \cite{farabet2012comparison}. The processing mechanism of SNN can be realized in neuromorphic hardware, such as Truenorth \cite{merolla2014million}, Loihi \cite{davies2018loihi}, Darwin \cite{ma2024darwin3}, etc.

The energy efficiency advantages of SNNs make them well-suited for deployment on low-power edge devices. These devices typically have significant resource constraints, such as limited memory capacity and computing resources. Yet, achieving high accuracy with SNN often requires large networks and extended processing timesteps, as they provide superior feature learning and more nuanced data representation. However, this approach increases the demands on memory and computation, which poses challenges for deployment on resource-constrained devices and undermines the potential advantages of SNNs.

Neural architecture search (NAS) aims to automate the neural network design under specific resource constraints. Recent studies have applied NAS to SNN to enable flexible and effective network design and have demonstrated better performance compared to manually designed SNN architectures. However, most of these works do not consider resource constraints in the search process \cite{na2022autosnn,che2022differentiable} and only aim at improving the accuracy. While Kim et al. consider the number of spikes, thereby reducing communication overhead \cite{kim2022neural}, this metric alone is insufficient for a comprehensive evaluation of resource consumption. 
To better evaluate resource utilization, we use model size for assessing memory footprint, which is also related to memory access energy, and computation complexity for measuring computational cost. These metrics offer more direct and comprehensive indicators of memory and computing resource demands \cite{zhang2016cambricon}.

To design a lightweight and efficient SNN characterized by small model size and low computation complexity as well as high performance, we incorporate compression within the NAS framework. Existing model compression techniques, including pruning and quantization, reduce the number of parameters and the number of bits used to represent each parameter, respectively, benefiting memory footprint, power efficiency and computational complexity.  However, existing SNN pruning and quantization works are all based on fixed architectures, leaving open the question of how to integrate them into evolving SNN architectures in search process. In addition to spatial domain, SNNs uniquely feature a temporal domain that decides time iterations (also known as timesteps).  By compressing these iterations, we can directly reduce computational operations and save energy. As current SNN-based NAS methods all originate from ANNs and lack temporal considerations, the potential of implementing the compressive timesteps in NAS has not yet been exploited. 

In this paper, we introduce a novel approach named LitE-SNN that incorporates both spatial and temporal compression into the automated network design process, aimed at designing SNNs with compact models and lower computational complexity. Spatially, we propose a Compressive Convolution block (CompConv) that expands the search space to incorporate pruning and quantization. Considering the variable synaptic formats in neuromorphic hardware, CompConv supports mixed-precision quantization to make the network design more flexible and efficient. Meanwhile, we utilize the shared weights and shared pruning masks to reduce the computation introduced by mixed-precision. 
Temporally, we propose a compressive timestep search that is tailored for the temporal domain of SNNs. Given the computational cost constraint, the model can find the optimal timesteps in accordance with the architecture and spatial compression.
Finally, we formulate a multi-objective joint optimization that simultaneously determines the network architecture and compression strategies. This joint optimization ensures a cohesive optimization of these factors, avoiding local optimal and suboptimal global solutions resulting from multiple single-objective optimizations.
We conduct the experiments on two image datasets CIFAR-10, CIFAR-100 and a speech dataset Google Speech Command.
Experimental results show that our proposed LitE-SNNs achieve competitive accuracy with remarkably smaller model sizes and fewer computation costs.
Furthermore, we validate the effectiveness of our LitE-SNNs in balancing accuracy with resource cost and the superiority of our joint optimization. Finally, we conduct energy analysis to further confirm the energy efficiency of LitE-SNNs.

% The main contributions can be summarized as follows:

% We propose an SNN-based NAS framework that incorporate the spatial compression (pruning and quantization) and temporal compression (time length).

\section{Related Work}
This section reviews current developments in neural architecture search (NAS) and model compression. NAS aims to automatically design neural architectures that achieve optimal performance using limited resources. Model compression aims to achieve lightweight neural networks with comparable performance, including pruning and quantization, etc.
\subsection{Neural Architecture Search}
%Neural architecture search (NAS) aims to automatically design neural architectures that achieve optimal performance using limited resources. 
In ANNs, NAS networks have surpassed manually designed architectures on many tasks such as image classification \cite{liu2019darts,cai2019proxylessnas}. In the SNN domain, \cite{na2022autosnn} presented a spike-aware NAS framework involving direct supernet training and an evolutionary search algorithm with spike-aware fitness. However, this approach is confined to exploring a few predefined blocks, limiting the potential to discover optimal designs beyond them.
\cite{kim2022neural} selected the architecture that can represent diverse spike activation patterns across different data samples. \cite{che2022differentiable} proposed a spike-based differentiable hierarchical search based on the DARTS \cite{liu2019darts} framework. 
The SNNs designed by these two works are more flexible and have better accuracy, however, their focus is solely on the accuracy aspect, neglecting critical factors like resource costs. Moreover, they do not consider the time domain during the search process, which is an important aspect of SNN.
%可支持的结构比较简单

\subsection{Model Compression}
Pruning involves removing unnecessary connections to reduce the number of parameters and computational needs. In ANNs, many existing studies employed important scores-based pruning \cite{sehwag2020hydra}, which involves training importance scores for each weight and masking less important weights. In SNNs, \cite{chen2021pruning} proposed a gradient-based rewiring method to learn the connectivity and weight parameters of SNNs. \cite{chen2022state} addressed the pruning of deep SNNs by modeling the state transition of dendritic spine. \cite{kim2022exploring} introduced the lottery ticket hypothesis-based pruning for sparse deep SNNs.

Quantization reduces the number of bits used to represent each parameter to reduce the memory footprint and computation complexity. Some ANN-based works \cite{cai2017deep,cai2020rethinking} set the quantizer based on the weight and activation distribution. \cite{rueckauer2017conversion} converted the binary ANN to obtain a binary SNN. \cite{srinivasan2019restocnet} used STDP to learn the one-bit weight synapses. However, these SNN methods apply uniform bit-width, neglecting the different sensitivities of different filters. \cite{lui2021hessian} employed a layer-wise Hessian trace analysis for mixed-precision quantization, allocating the bit-width per layer, but limited to a three-layer shallow network.
%bit-width allocation per layer, but it is limited to a three-layer shallow network.

There are several works that consider both pruning and quantization. \cite{deng2021comprehensive} adapted alternating direction method of multipliers (ADMM) optimization with spatio-temporal backpropagation for SNN pruning and quantization. However, this approach is performed sequentially and may lead to suboptimal results, e.g., the best network architecture for the dense and full-precision model is not necessarily the optimal one after pruning and quantization \cite{wang2020apq}. \cite{rathi2018stdp} presented a joint pruned and quantized SNN with STDP-based learning, but it is limited to a fixed and shallow (two layers) structure. 
Current SNN research lacks a comprehensive approach that simultaneously considers the neural architecture design, pruning and quantization. Therefore, we need a solution to jointly optimize these aspects. Similar work in ANN needs to train a large supernet and distill a number of smaller sub-networks, followed by training the accuracy predictors and evolutionary search \cite{wang2020apq}. The whole training process takes 100 GPU days on V100 GPU. Considering that SNNs usually require more training time than ANNs \cite{kim2022neural}, it is not practical to utilize this approach for SNNs. 

In the time domain, \cite{li2023input} dynamically determines the number of timesteps during inference on an input-dependent basis. While this work effectively reduces timesteps, it is limited to the inference stage and cannot be optimized together with other training parameters, which also makes it unsuitable for use in NAS. In this paper, we will jointly optimize the pruning, mixed-precision quantization, and timestep compression through the NAS framework.

\section{Method}
In this section, we first introduce the preliminary knowledge of spiking neurons and architecture search space, then present our proposed spatial and temporal compression search, and finally explain the spatial-temporal joint optimization.
\subsection{Preliminary}
%spiking neuron integrates input spikes from the previous layer into membrane potential. When the membrane potential integrated over time exceeds a certain threshold, the neuron will fire a spike to the next layer. 
%autosnn说法
\begin{figure}[!b]
	\centering
	%\fbox{
		\includegraphics[width=1.0\linewidth]{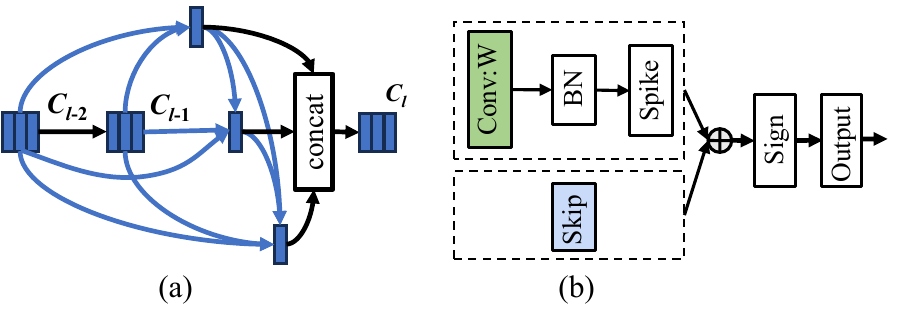}
		
	%}
	\caption{Hierarchical search space. (a) Cell structure search space with 3 nodes. (b) Within one node, two candidate operations are summed filtered by a sign function.}
	\label{fig:spikedhs}
\end{figure}
We choose a widely used spiking neural model Leaky Integrate-and-Fire (LIF) \cite{zheng2021going} to describe the neural behavior, which can be written as: 
\begin{equation}\label{fneuron1}
	u_i^{t,n} = \tau_{decay} u_i^{t-1,n} (1-y_i^{t-1,n}) + \sum_{j}W_{i,j}^{n} y_j^{t, n-1}
\end{equation}
where $\tau_{decay}$ denotes the membrane decay constant, $W_{ij}$ is the synaptic weight between the $j$-th neuron (or feature map) and the $i$-th neuron (or feature map). If the membrane potential at timestamp $t$ of the neuron in $n$-th layer $u^{t,n}$ is larger than the threshold $V_{th}$, the output spike $y_i^{t,n}$ is set to 1; otherwise it is set to 0. 
%cell的关系
 Our method builds upon the search space of spikeDHS \cite{che2022differentiable}, which is a spiking-based differentiable hierarchical search framework that can find best-performed architectures for SNNs.
 The search space consists of a number of cells and each cell (depicted in Figure \ref{fig:spikedhs}(a)) is defined as a repeated and searchable unit with N nodes (depicted in Figure \ref{fig:spikedhs}(b)), $\{x_i\}_N$. Each cell receives input from two previous cells and forms its output by concatenating all outputs of its nodes. Each node can be described by:
\begin{equation}
    x_j = f(\sum_{i<j} o^{i,j}(x_i))
\end{equation}
where $f$ is a spiking neuron taking the sum of all operations as input, $o^{(i,j)}$ is the operation associated with the directed edge connecting node $i$ and $j$. During search, each edge is represented by a weighted average of candidate operations. The information flow connecting node $i$ and node $j$ becomes:
\begin{equation}\label{eq3}
    \bar{o}^{(i,j)} (x) = \sum_{o\in \mathcal{O}^{i,j}}\frac{exp\big(\alpha_o^{(i,j)}\big)}{\sum_{o\in \mathcal{O}^{i,j}}exp \big(\alpha_o^{(i,j)}\big)}\cdot o (x)
\end{equation}
where $\mathcal{O}^{i,j}$ denotes the operation space on edge $(i, j)$ and $\alpha_o^{(i,j)}$ is the weight of operation $o$, which is a trainable continuous variable. After search, a discrete architecture is selected by replacing each mixed operation $\bar{o}^{(i,j)}$ with the most likely operation $o^{i,j}$ that has $\mathop{max}\limits_{o\in \mathcal{O}^{(i,j)}}\alpha_o^{(i,j)}$.
%SpikeDHS optimizes weight parameters ($w$) and architecture parameters ($\alpha$) with the following bi-level optimization: 1) update weights w by $\nabla_w L(w,\alpha)$ and 2)update architecture parameters $\alpha$ by $\nabla_\alpha L(w,\alpha)$.
%compconv能不能按这个形式
 
\subsection{Compressive Convolution (CompConv)}
\begin{figure*}[!t]
	\centering
	%\fbox{
		\includegraphics[width=1.0\linewidth]{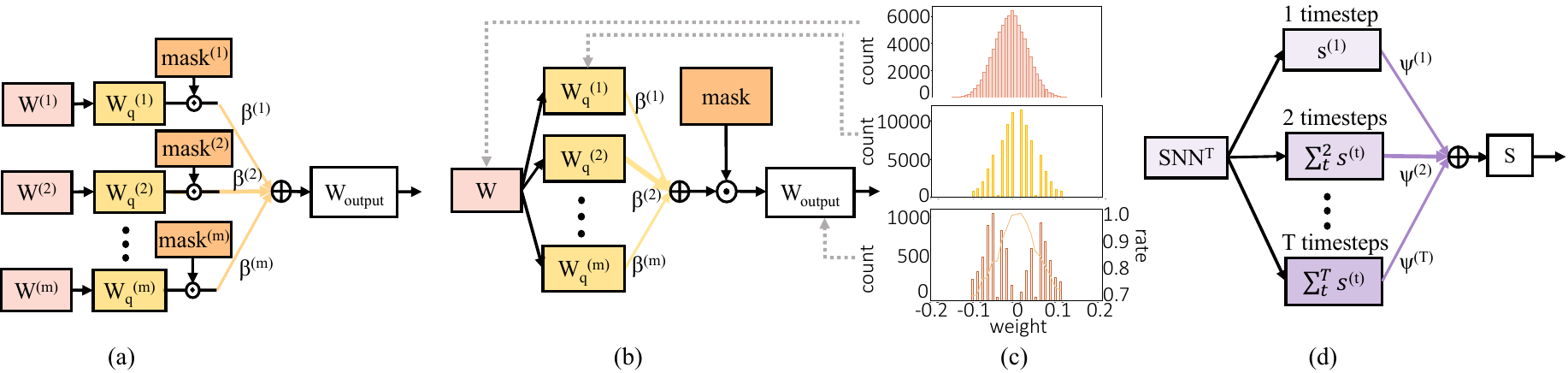}
		
	%}
	\caption{(a) A naive solution that integrates mixed-quantization with pruning. (b) Our proposed CompConv for a more efficient integration. (c) Visualization of the distribution of $W$, $\bar e(W)$, and $W_{output}$ with pruning rates. (d)  Our proposed timestep search solution. }
	\label{fig:cconv}
\end{figure*} 
To design lightweight and efficient SNNs, in this work, we incorporate compression techniques into the neural architecture search to take advantage of the efficiency and flexibility of automatic network design.
To this end, we propose a novel compressive convolutional block (CompConv) to replace the standard convolution (green block in Figure \ref{fig:spikedhs}) to enable the search of compressive SNNs. Neuromorphic hardware like Loihi is capable of supporting sparse kernels and variable synaptic formats \cite{davies2018loihi,davies2021taking}, which inspired us to utilize weight pruning and mixed-precision quantization for compression. 

%  To facilitate network weight pruning, we employ the pruning mask based on the importance score $s$ assigned to each weight. $s$ is set as learnable parameters and initialized based on the magnitude of the weights \cite{mousavi2022pr}:
% \begin{equation}
%     s_i = \frac{1}{max(|w|)} * w_i
% \end{equation}
% where $w_i$ represents the $i$-th convolution weight and $s_i$ denotes its corresponding importance.
% With a pruning rate of $p\%$, we retain the weights with the top $1-p\%$ importance and the mask can be obtained by
% \begin{equation}\label{fneuron2}
% 	mask_i = 
% 	\left\{ \begin{array}{@{}ll@{}}
% 		1, & \text{if} \ s_i \in top \ 1-p\%\\
% 		0,
% 		& \text{otherwise}\\
% 	\end{array}\right.
% \end{equation}

% The pruning result are represented by the mask, which is a binary matrix where each element indicates whether the weights have been pruned (assigned a 0) or retained (assigned a 1) of $j$-th branch.

As shown in Figure \ref{fig:cconv}(a), \textbf{a naive solution} to integrate mixed-precision quantization with pruning is to organize each bit-width quantization operation and the corresponding pruning mask into a separate branch and update each branch individually, which can be described 
\begin{equation}\label{eq:sf1}
   W^{(j)}_{q} = Q_j(W^{(j)})
\end{equation}
\begin{equation}\label{eq:sf2}
 e^{(j)}(W) = W_q^{(j)} \odot mask^{(j)} 
\end{equation}
\begin{equation}\label{eq:sf3}
\bar{e} (W) =  \sum_{j}^{len(B)} \frac{exp\big(\beta^{(j)}\big)}{\sum_j^{len(B)} exp \big(\beta^{(j)}\big)} e^{(j)}(W)
\end{equation}
\begin{equation}\label{eq:sf4}
   W_{output}  =  \bar{e} (W) 
\end{equation}
where $W^{(j)}$ represents the weight tensor of $j$-th branch, 
$Q_j$ is a quantization function that follows \cite{cai2017deep}, which quantizes the input value by the $j$-th candidate bitwidth. $mask^{(j)} $ is a binary matrix where each element indicates whether the weights have been pruned (assigned a 0) or retained (assigned a 1) of $j$-th branch. The pruning method follows the \cite{sehwag2020hydra}.
$W_q^{(j)}$ and $ e^{(j)}(W)$ represent the quantized weights and operation of $j$-th branch respectively. 
$\beta^{(j)}$ is the weight of $j$-th operation and $B$ denotes the bit-width candidate space.
However, this approach significantly increases the computation, as we need to update the weights and mask for each individual branch. Additionally, it has been found in \cite{cai2020rethinking} that 
branches with low $\beta$ values receive few gradients, potentially leading to the under-trained of weights and mask.
% \begin{figure}[!t]
% 	\centering
% 	%\fbox{
% 		\includegraphics[width=1.0\linewidth]{distribution.pdf}
		
% 	%}
% 	\caption{Our proposed compressive convolution block (CompConv) which effectively incorporates the mix-precision quantization and pruning into the search space.}
% 	\label{fig:distribution}
% \end{figure} 
In contrast, our CompConv provides \textbf{a more efficient solution} by sharing the weights and pruning mask among all branches within a block. As depicted in Figure \ref{fig:cconv}(b), we first quantize $W$ with different bit-width in different branches to get $W_q$ and then perform pruning $mask$ on the weighted average of $W_q$. The Equation (\ref{eq:sf1}-\ref{eq:sf4}) can be rewritten as
\begin{equation}\label{eq:e1}
   W^{(j)}_{q} = Q_j(W)
\end{equation}
\begin{equation}\label{eq:e2}
 e^{(j)}(W) = W_q^{(j)} 
\end{equation}
\begin{equation}\label{eq:e3}
\bar{e} (W) =  \sum_{j}^{len(B)} \frac{exp\big(\beta^{(j)}\big)}{\sum_j^{len(B)} exp \big(\beta^{(j)}\big)} e^{(j)}(W)
\end{equation}
\begin{equation}\label{eq:quantize4}
   W_{output} = \bar{e} (W) \odot mask
\end{equation}
where we remove the $W^{(j)}$ and $mask^{(j)}$ in each branch and utilize universal shared $W$ and $mask$ instead.
This modification allows the gradient to be fully applied to the  $W$ and $mask$, eliminating the possibility of underfitting and simplifying the search space. Overall, our proposed CompConv offers a more efficient solution for achieving compression with mixed-precision quantization and pruning in the search space of automatic lightweight SNN designs.

 Figure \ref{fig:cconv}(c) illustrates the distributions of $W$, $\bar e(W)$, and $W_{output}$ with pruning. $W$ displays a symmetric, continuous distribution with a diminishing central peak. $\bar e(W)$, quantized at 2-bit and 4-bit precision, follows a similar discrete pattern. The pruning rate curve indicates that weights with smaller absolute values are pruned more for their lower importance, resulting in $W_{output}$ having higher values on the sides and lower at the center and far sides.
 %The pruning rate curve indicates that weights with smaller absolute values undergo more pruning due to their lower importance.  Thus, the distribution of the $W_{output}$ is high on the sides and low at the center and far sides.
\subsection{Compressive Timestep Search}
%Our LitE-SNN also extends the search space to the time domain to incorporate the  timesteps into the automatic design process. Timesteps is a significant factor for SNN to reduce the number of computational operations, thereby saving energy and decreasing latency \cite{li2023input}. As the network tends to use long timesteps for high accuracy, our timestep search need to be conducted under computation cost constraints. We define the computation cost utilizing the widely-accepted Synaptic Operations (SynOps) metric:
The selection of timesteps in SNNs is crucial, as it influences the network's ability to process temporal information. A longer timestep can improve accuracy but comes with increased computational cost due to the necessity for more time iterations. Conversely, a shorter timestep may reduce computational cost but potentially at the expense of temporal detail and accuracy. Recognizing this, our work focuses on incorporating the timestep selection into the search process to automatically balance computational cost with performance.

To assess the computational operations associated with different timesteps, 
we follow the convention of the neuromorphic computing community to use the total synaptic operations (SynOps) metric \cite{wu2021tandem}, defined as: 
 \begin{equation}
    SynOps = (1-p\%) k_h  k_w  C_{in}  C_{out} H  W S
\end{equation}
where $p$ is the pruning rate, $k_h$ and $k_w$ are the kernel height and width, $C_{in}$ and $C_{out}$ the input and output channel sizes, $H$ and $W$ the height and width of the feature map. $S$ is the average spike rate, which is originally defined as
 \begin{equation}\label{eq:spikerate}
	S = \sum_t^{T}s^{(t)}
\end{equation}
where $T$ is the total time length, $s^{(t)}$ indicates the average spike rate per neuron at the $t$-th timestep. We observe that the number of timesteps affects the spike rate $S$ and, consequently, the network's computational cost. As shown in  Figure \ref{fig:cconv}(d), to enable this search, we consider a range of candidate timestep counts $\{1, 2, \ldots, T\}$ and modify the Equation (\ref{eq:spikerate}) by introducing trainable weights for each timestep count:
 \begin{equation}\label{eq:timesearch}
    S = \sum_t^{T} \left(\frac{exp\big(\psi^{(t)}\big)}{\sum_t^T exp \big(\psi^{(t)}\big)} \sum_{t'}^{t} s^{(t')}\right)
\end{equation}
where $\psi^{(t)}$ is the weight of having $t$ timesteps, $\frac{exp(\psi^{(t)})}{\sum_t^T exp (\psi^{(t)})}$ is the relative weight with the total sum equaling $1$.  
%For illustration, consider a scenario where the total time length $T$ is 6, and the relative weights for each timestep $t \in \{1, 2, 3, 4, 5, 6\}$ are $[0, 0, 0, 0, 0.6, 0.4]$. The spike rate $S$ is calculated as the sum of spike rates for the first 5 timesteps, plus 40\% of the spike rate from the $6$-th timestep.
%If the weight distribution changes to $[0, 0, 0, 0, 0, 1]$, the spike rate $S$ is computed as the sum of spike rates across all 6 timesteps, which degenerate to Equation (\ref{eq:spikerate}). The trained weights $\psi$ allow us to evaluate the impact of the timestep selection on the overall network performance. We will introduce how to balance network performance against computational cost in Section \ref{loss}.
For illustration, consider a scenario where the total time length $T$ is 6 with relative weights for each timestep $t \in \{1, 2, 3, 4, 5, 6\}$ as $[0, 0, 0, 0, 0.6, 0.4]$. Here, the spike rate $S$ includes the sum of spike rates for the first 5 timesteps plus 40\% from the 6th. If weights change to $[0, 0, 0, 0, 0, 1]$, $S$ equals the sum across all timesteps, matching Equation (\ref{eq:spikerate}). The trained weights $\psi$ evaluate the impact of timestep selection on network performance. We will introduce how to balance network performance against computational cost in Section \ref{loss}

\subsection{Loss Function and Joint Optimization Algorithm}\label{loss}
The objective of LitE-SNN is to identify SNNs that are both lightweight and efficient, achieving high performance while minimizing memory and computation costs. This can naturally be formulated as a multi-objective optimization problem. We formulate the following loss function:
\begin{equation} \label{equ:loss}
    Loss = Loss_{CE} + \lambda_1 Loss_{MEM} + \lambda_2 Loss_{COMP}
\end{equation}
 where the coefficients $\lambda_1$ and $\lambda_2$ serve as regularization parameters that control the trade-off among  $Loss_{CE}$, $Loss_{MEM}$ and  $Loss_{COMP}$. 
The first term $Loss_{CE}$ represents the cross-entropy (CE) loss, responsible for evaluating the predictive accuracy of the SNN. Following  \cite{che2022differentiable}, the original definition of $Loss_{CE}$ is:
\begin{equation}
	Loss_{CE} =  CE^{(T)} =  CE\left(\frac{1}{T} \sum_{t=1}^{T} O(t), y\right),
\end{equation}
where $O(t)$ denotes the output of the SNN and $y$ represents the target label.
But as $Loss_{CE}$ is also influenced by the number of timesteps, we rewrite it as:
\begin{equation}
    Loss_{CE} = \sum_t^{T}\frac{exp\big(\psi^{(t)}\big)}{\sum_t^T exp \big(\psi^{(t)}\big)}\cdot CE^{(t)}
\end{equation}
The second term $Loss_{MEM}$ quantifies the memory usage of the network, which is measured by the model size:
	\begin{align}
		Loss_{MEM} &= b_w k_h  k_w  C_{in}  C_{out} (1-p\%)
	\end{align}
 where $b_w$ is the bit-width of the network weights. During search, the $b_w$ is a weighted average of candidate bit-width and can be calculated as 
 \begin{equation}
    b_w = \sum_j^{len(B)}\frac{exp\big(\beta^{(j)}\big)}{\sum_j^{len(B)} exp \big(\beta^{(j)}\big)}\cdot q^{(j)}
\end{equation}
where $q^j$ denotes the $j$-th bitwidth candidate. 
 The third term $Loss_{COMP}$ assesses the computational cost, which is affected not only by the SynOps but also by the operational precision \cite{cai2020rethinking}. Therefore, we define this term as bit-SynOps, reflecting the computational complexity of the model.
 The corresponding equation is:
 \begin{equation}
 	Loss_{COMP} = Bit\text{-}SynOps = b_w SynOps
 \end{equation}

The joint optimization comprises two steps for each iteration. First, we conduct the first forward pass and update the weights, which constitute the foundational components of the network. Then, a second forward pass is conducted using the updated weights and we simultaneously update $\beta$ and $mask$ within our CompConv, as well as the $\psi$ and architecture parameters $\alpha$. Both two steps apply the loss function in Equation (\ref{equ:loss}). Our joint optimization ensures cohesive and efficient optimization, avoiding local optimal and suboptimal global solutions resulting from multiple single-objective optimizations \cite{wang2020apq}.

After search, we decode the cell structure (in Figure \ref{fig:spikedhs}(a)) by retaining the two strongest incoming edges for each node and select the node operation (in Figure \ref{fig:spikedhs}(b)) with the strongest edge, where the strength of edge is determined by $\alpha$. 
The bit-width (in Figure \ref{fig:cconv}(b)) chosen for each cell is decoded by the branch with the highest weight $\beta$. 
We also select the number of timesteps with the highest weight $\psi$ (in Figure \ref{fig:cconv}(d)). 
During the retraining, we use the auxiliary loss as in \cite{che2022differentiable}. The weights and masks are retrained to adapt to single-branch network. We adopt the surrogate gradient to direct train the SNN and use the Dspike function \cite{li2021differentiable} to approximate the derivative of spike activity.

\section{Experiments}
This section compares our model's performance and efficiency with current benchmarks and optimization methods and evaluates the effectiveness of parameters $\lambda_1$ and $\lambda_2$.

\subsection{Experimental Settings}
\noindent \textbf{Datasets.}
We evaluate our proposed LitE-SNN on two image datasets: CIFAR-10 \cite{krizhevsky2009learning}, CIFAR-100 \cite{krizhevsky2009learning}. %and an audio dataset: Google Speech Command \cite{warden2018speech}.
We also choose an audio dataset Google Speech Command (GSC) \cite{warden2018speech} which represents the keyword-spotting task commonly deployed on edge devices for quick responses.

CIFAR-10 has 60,000 images of 10 classes with a size of 32$\times$32. Among them, there are 50,000 training images and 10,000 testing images. CIFAR-100 has the same configurations as CIFAR-10, except it contains 100 classes. We split the dataset into 9,000 training samples and 1,000 test samples. The pre-processing method we use on these two datasets is the same as \cite{che2022differentiable}.

GSC has 35 words spoken by 2,618 speakers. Following the pre-processing method used in \cite{yang2022deep}, we split the dataset into 12 classes, which include 10 keywords and two additional classes. We apply Mel Frequency Cepstrum Coefficient (MFCC) to extract acoustic features. The sampling frequency is $16$kHz, the frame length and shift are set to $30$ms and $10$ms, and the filter channel is defined as $40$.

\noindent \textbf{Hyperparameters.}
We implement our LitE-SNN using Pytorch on NVIDIA GeForce RTX 3080 (10G) GPUs. During the architecture search, we conduct $50$ epochs with a batch size of $30$. We use the SGD optimizer with momentum $0.9$ and a learning rate of $0.025$ to update network weights $w$ and employ the Adam optimizer with a learning rate of $3e^{-4}$ to update $\beta$, $mask$, and $\alpha$. During the retraining, we train for $200$ epochs with a batch size of $50$, using the SGD optimizer with momentum $0.9$ and a cosine learning rate of $0.025$. The neuron parameters $\tau_{decay}$ and $V_{reset}$ are set to $0.2$ and $0$ respectively. The threshold $V_{th}$ is initialized to $0.5$. The temperature parameter in Dspike function is set to $3$ and the settings of other parameters follow \cite{che2022differentiable}.

\begin{table}[!t]
	\centering
	\resizebox{0.8\linewidth}{!}{
		\begin{tabular}{c|cc}
			\toprule
			& \multicolumn{1}{c}{CIFAR} & \multicolumn{1}{c}{GSC} \\
			\multirow{1}{*}{Layer} & \multicolumn{1}{c}{$\text{channel} \times \text{feature size}$} & \multicolumn{1}{c}{$\text{channel} \times \text{feature size}$} \\
			\midrule
			stem & $(C_{\text{CIFAR}}) \times (32 \times 32)$ & $(C_{\text{GSC}}) \times (40 \times 98)$ \\
			\midrule
			Cell1 & $(C_{\text{CIFAR}}) \times (32 \times 32)$ & $(C_{\text{GSC}}) \times (40 \times 98)$ \\
			Cell2 & $(C_{\text{CIFAR}}) \times (32 \times 32)$ & $(C_{\text{GSC}}) \times (40 \times 98)$ \\
			Cell3 & $(2 \times C_{\text{CIFAR}}) \times (16 \times 16)$ & $(2 \times C_{\text{GSC}}) \times (20 \times 49)$ \\
			Cell4 & $(2 \times C_{\text{CIFAR}}) \times (16 \times 16)$ & $(2 \times C_{\text{GSC}}) \times (20 \times 49)$ \\
			Cell5 & $(2 \times C_{\text{CIFAR}}) \times (16 \times 16)$ & $(4 \times C_{\text{GSC}}) \times (10 \times 25)$ \\
			Cell6 & $(4 \times C_{\text{CIFAR}}) \times (8 \times 8)$ & $(4 \times C_{\text{GSC}}) \times (10 \times 25)$ \\
			Cell7 & $(4 \times C_{\text{CIFAR}}) \times (8 \times 8)$ & - \\
			Cell8 & $(4 \times C_{\text{CIFAR}}) \times (8 \times 8)$ & - \\
			\midrule
			Pooling & $(4 \times C_{\text{CIFAR}}) \times (1 \times 1)$ & $(4 \times C_{\text{GSC}}) \times (1 \times 1)$ \\
			FC & 10 / 100 & 12 \\
			\bottomrule
	\end{tabular}}
	\caption{Network backbone for CIFAR and GSC datasets.}
	\label{parameter}
\end{table}

\begin{table*}[!t]
\centering
\resizebox{1.0\linewidth}{!}{
\begin{tabular}{lcccccccccc}
\toprule
 & \multicolumn{5}{c}{\multirow{2}{*}{\textbf{CIFAR-10}}} & \multicolumn{5}{c}{\multirow{2}{*}{\textbf{CIFAR-100}}}\\
 \\
 \midrule
\multirow{2}{*}{\textbf{Model}} & \textbf{Acc.} & \textbf{Model Size} & \textbf{Bit-SynOps} & \textbf{\textbf{Bitwidth.}} & \#\textbf{Timesteps} & \textbf{Acc.} & \textbf{Model Size} & \textbf{Bit-SynOps} & \textbf{\textbf{Bitwidth.}} & \#\textbf{Timesteps}\\
& (\%) & {(MB)} & (M) & (b) & & (\%) & {(MB)} & (M) & (b) &\\
\midrule
\cite{rueckauer2017conversion} & 83.35 & 0.27 & - & 1 &4 & \multicolumn{5}{c}{-} \\
	\midrule
\cite{srinivasan2019restocnet} & 66.23 & 2.10 & - & 1 &25 & \multicolumn{5}{c}{-} \\
	\midrule
\multirow{2}{*}{\cite{chen2021pruning}} & 92.54 & 41.73 & 847 & 32 (FP) &8 & 69.36 & 11.60 & 1179 &32 (FP) &8 \\
& 92.50 & 17.68 & 638 &32 (FP) &8 & 67.47 & 2.56 & 882 & 32 (FP) &8  \\
	\midrule
\multirow{2}{*}{\cite{deng2021comprehensive}} & 87.84 & 8.62 & - & 3 &8 & 57.83  & 5.75 & - & 3 &8  \\
& 87.59 & 2.87 & - & 3 &8  & 55.95 & 2.87 & - & 1 &8  \\
	\midrule
\multirow{2}{*}{\cite{kim2022exploring}} & 93.50 & 12.02 & 950 &32 (FP) &5  & 71.45 & 12.02 & 663 & 32 (FP) &5  \\
& 93.46 & 2.88 & 560 & 32 (FP) &5  & 71.00 & 2.88 & 517 & 32 (FP) &5  \\
	\midrule
\multirow{2}{*}{\cite{chen2022state}} & 92.49 & 3.28 & 1290 & 32 (FP) &8  & \multicolumn{5}{c}{\multirow{2}{*}{-}} \\
& 90.21 & 1.10 & 1245 & 32 (FP)  &8  \\
	\midrule
\multirow{2}{*}{\cite{kim2022neural}} & 94.12 & 199.72 & 16504 & 32 (FP) &8   & 73.04 & 82.84 & 39243 & 32 (FP) &5   \\
& 93.73 & 168.24 & 13203 & 32 (FP) &5  & 70.06 & 77.60 & 27735 & 32 (FP) & 5  \\
	\midrule
\multirow{2}{*}{\cite{na2022autosnn}} & 93.15 & 83.68 & 18739& 32 (FP) &8   & \multirow{2}{*}{69.16} & \multirow{2}{*}{42.76} & \multirow{2}{*}{19547} & \multirow{2}{*}{32 (FP)} & \multirow{2}{*}{8}\\
& 92.54 & 21.76 & 12787 & 32 (FP) &8  \\
	\midrule
\multirow{2}{*}{\cite{che2022differentiable}} & 95.50 & 56.00 & 41781 & 32 (FP) &6  & 76.25 & 56.00 & 51577& 32 (FP) &6  \\
& 95.36 & 56.00 & 40564 & 32 (FP) &6  & 76.03 & 48.00 & 50270& 32 (FP) &6  \\
	\midrule
\multirow{1}{*}{\cite{xu2023constructing}} & 93.41 & 44.68 & - & 32 (FP) &4  & \multicolumn{5}{c}{-} \\
	\midrule
\cite{shen2023esl} & 91.09 & 25.12 & - & 32 (FP) &2  & 73.48 & 25.12 &- & 32 (FP) &4 \\
\midrule
\multirow{2}{*}{\cite{yan2024efficient}} & 94.64 &  93.88 & 13184 & 32 (FP) & 3  & 74.78 & 110.20 & 16544 & 32 (FP) &3   \\
& 94.27 & 64.72 & 8640 & 32 (FP) &3  & 73.21 & 69.28 & 11136 & 32 (FP) & 3  \\
\midrule
Our Work (large) &  95.60$\pm$0.24 &3.60  & 2863 & searched &searched (6) & 77.10$\pm$0.04 &3.62 & 3590 & searched & searched (6) \\
Our Work (medium) &94.52$\pm$0.05  & 1.23 & 913  & searched & searched (5)   &74.51$\pm$0.35  &1.27 &1110 & searched &searched (5) \\
Our Work (small) &91.98$\pm$0.15 &0.55 & 298 &searched &searched (4)  & 69.55$\pm$0.11 &0.59  & 372 & searched &searched (4) \\
\bottomrule
\end{tabular}}
\caption{Performance comparison with the state-of-the-art lightweight SNNs on CIFAR10 and CIFAR100 datasets.}
\label{cifar}
\end{table*}

\noindent \textbf{Architecture.}
The architecture backbone is reported in Table \ref{parameter}. The stem layer has a structure of Conv-Spike-BN with a kernel size of $3 \times 3$. The initial channel counts are denoted as $C_{\text{CIFAR}}$ and $C_{\text{GSC}}$ for CIFAR and GSC datasets, with respective values of $48$ and $16$ during the search.
For the rest of the network, we employ 8 searchable cells on CIFAR dataset and 6 searchable cells on GSC dataset. The final two layers are a global pooling and a fully-connected layer.
Each of the searchable cells has 4 nodes with each node containing two candidate operations: Conv 3$\times$3 and skip. 
In this paper, the bit-width candidate space is set as \{1, 2, 4\}. 
For image datasets, we search on CIFAR-10 and then retrain on target datasets including CIFAR-10 and CIFAR-100. 
For audio dataset GSC, we use the architecture searched on its own.
During retraining, $C_{\text{CIFAR}}$ is increased to $144$ to extract more features from raw data.
To compare with state-of-the-art SNNs which span a wide range of model size and Bit-SynOps on CIFAR, we set three scales of LitE-SNNs. Large scale uses the $\lambda_1$, $\lambda_2$ and $p$ of $0$, $0$ and $50$; medium scale employs $5e^{-10}$,  $5e^{-14}$ and $80$; and small scale adopt $1e^{-9}$,  $1e^{-13}$ and $90$. On GSC, we use $\lambda_1$, $\lambda_2$ and $p$ of $1e^{-9}$, $1e^{-13}$ and $70$.

\subsection{Comparison with Existing SNNs}
Table \ref{cifar}\footnote{The results of comparison SNNs are from original papers, or (if not provided) from our experiments using publicly available code from the paper. If the original paper reported multiple results, we list two results with the highest accuracy in the table.} presents a comprehensive comparison between our proposed LitE-SNNs and other state-of-the-art lightweight SNNs on CIFAR10 and CIFAR-100 datasets. 
Our large-scale LitE-SNN achieves state-of-the-art accuracy on CIFAR-10 and CIFAR-100 datasets while maintaining a remarkably compact model size of 3.60MB and 3.62MB and relatively fewer BitSynOps of 2863M and 3590M. We can see that, this compressed SNN achieves superior accuracy compared with the baseline model without compression \cite{che2022differentiable}. This result reveals the potential of compression to remove noisy and irrelevant connections to enhance the ability to capture valuable information.
Our small-scale LitE-SNN has a model size of only 0.55MB and Bit-SynOps of 298M, smaller than all other models except the model size of \cite{rueckauer2017conversion} on CIFAR-10 dataset. Nevertheless, it achieves a competitive accuracy of 91.98\%, outperforming \cite{rueckauer2017conversion} by a big margin of 8.63\%. 
Notice that \cite{chen2021pruning}, \cite{kim2022exploring} and \cite{chen2022state} also maintain a good performance with relatively small model size and Bit-SynOps. This is because they employ effective pruning methods to reduce the number of parameters. Despite this, under similar model sizes or Bit-SynOps, our LitE-SNNs still achieve higher accuracy. For example, on CIFAR-10 dataset, our medium-scale LitE-SNN achieves 94.52\% accuracy with a 1.23MB model, surpassing \cite{kim2022exploring} by 1\% while being 1.6MB smaller.  
The results of our timestep search are also shown in Table \ref{cifar}. From large-scale to small-scale LitE-SNN, we observe that the optimal number of timesteps decreases from 6 to 4. This is because the increased $\lambda_1$ and $\lambda_2$ values in the search impose stricter resource limits. Consequently, the network tends to select a smaller number of timesteps to minimize calculations.
\begin{table}[!t]
\centering
\resizebox{1.0\columnwidth}{!}{
\begin{tabular}{lccccc}
\toprule
\multirow{2}{*}{\textbf{Model}} & \textbf{Acc.} & \textbf{Model Size} &\textbf{Bit-SynOps} & \textbf{Bitwidth.} & \textbf{\#\textbf{Timesteps}} \\
& (\%) & {(MB)} & (M) & (b) \\
\midrule
\cite{yilmaz2020deep} & 75.20 & 0.47 & -&32 (FP) & 10 \\
\cite{orchard2021efficient} & 88.97 & 0.95 &-& 32 (FP) & -  \\
\cite{yang2022deep} & 92.90 & 0.36 & 768&32 (FP) & 32 \\
\midrule
Our Work & 94.14$\pm$0.06 & 0.30 &37& searched & searched (3) \\
\bottomrule
\end{tabular}}
\caption{Performance comparison with other SNNs on Google Speech Command dataset.}
\label{GSC}
% \vspace{-0.35cm}
\end{table}

On the GSC dataset, as reported in Table \ref{GSC}, our model gets an accuracy of 94.14\% with a compact model size of 0.3M and Bit-SynOps of only 37M, which outperforms all the comparison methods in terms of accuracy and resource cost. It is worth noting that the model sizes of other compared methods are all less than 1MB. This is because GSC belongs to the keyword-spotting task that tends to run on the portable device. For such tasks, it is important to employ smaller models to minimize power consumption and speed up the response.

\subsection{Effectiveness of $\lambda_1$ and $\lambda_2$}
We vary $\lambda_1$ and $\lambda_2$ in Equation (\ref{equ:loss}) and validate its effect on the trade-off between accuracy and resource costs in the LitE-SNN on CIFAR-10 dataset. The other parameters maintain the same setting as our small-scale LitE-SNN.

\begin{figure}[!t]
	\centering
	%\fbox{
		\includegraphics[width=1.0\linewidth]{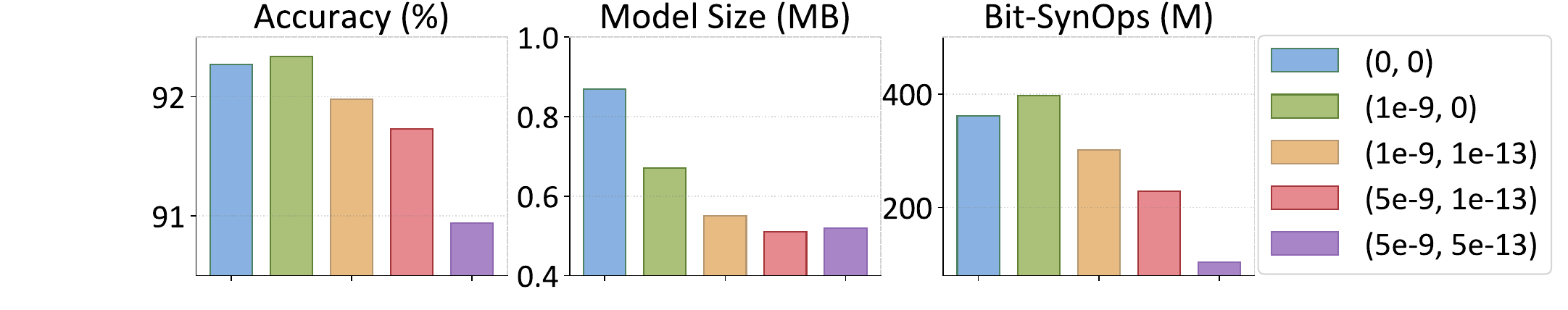}
		
	%}
	\caption{Ablation study of $\lambda_1$ and $\lambda_2$.}
	\label{fig:lambda}
\end{figure}

Figure \ref{fig:lambda} illustrates that 
increasing $\lambda_1$ and $\lambda_2$ leads to stricter resource limits, 
resulting in a trend of smaller network sizes and fewer Bit-SynOps, but at the expense of accuracy. Notably, $\lambda_1$ (the memory coefficient) and $\lambda_2$ (the computational cost coefficient) have interconnected effects on model size and Bit-SynOps. For instance, a rise in $\lambda_2$ alone, altering the network's search setting from (1e-9, 0) to (1e-9, 1e-13), impacts both Bit-SynOps and model size. This is because the constraint on Bit-SynOps limits model size, reducing both metrics. Similarly, adjusting the settings from (1e-9, 1e-13) to (5e-9, 1e-13) also decreases both model size and Bit-SynOps. The search results demonstrate that $\lambda_1$ and $\lambda_2$ can trade-off between accuracy and computational cost, while also coordinating changes in model size and Bit-SynOps.
% \begin{table}[!t]
% 	%\setlength\tabcolsep{2pt}  
% 	\begin{center}
%   \resizebox{1.0\columnwidth}{!}{
% 		\begin{tabular}{cccccc}
% 			\toprule
% 			 \textbf{$\lambda_1$}&\textbf{$\lambda_2$}&\textbf{Acc.(\%)} &\textbf{Model Size (MB)}  & \textbf{Bit-SynOps (M)}  & \#\textbf{Timesteps}  \\
%       \midrule
%        0& 0 & 92.27 & 0.87 & 361 & 3 \\
%     1e-9& 0 & 92.34 & 0.67 & 397 & 4 \\
%     1e-9& 1e-13 &  91.98 & 0.55 & 301 & 4\\
%     5e-9& 1e-13 &  91.73 & 0.51 & 229 & 4\\
%     5e-9& 5e-13 &  90.94 & 0.52 & 104 & 2\\
% 			\bottomrule
% 		\end{tabular}}
% 		\caption{Ablation study of $\lambda_1$ and $\lambda_2$.\qh{changing to fig}}
% 		\label{ablation:lambda} 
% 	\end{center}
%  \end{table}

\subsection{Comparison with Sequential Optimization}
\begin{table}[!t]
	\begin{center}
   \resizebox{0.9\columnwidth}{!}{
		\begin{tabular}{ccccc}
			\toprule
   \multirow{2}{*}{\textbf{Model}}& \textbf{Acc.} &\textbf{Model Size}&\textbf{Bit-SynOps}&\textbf{Design Cost}\\
   & (\%) &(MB)&(M)&(GPU days)\\
     \midrule
			 \multicolumn{5}{c}{Sequential Optimization}\\
      \midrule
         A + P + Q + T (large) & 94.73 & 3.43 &2426 & 7.04\\         
    A + P + Q + T (medium) &92.76 &1.40 &1001 & 7.04\\
     A + P + Q + T (small) &85.99 &0.73  &505 &7.04 \\
         % \cite{kim2022exploring} + Q &92.66 &1.52 &649 &2.33 \\
      \midrule
     	\multicolumn{5}{c}{Joint Optimization}\\
       \midrule
        LitE-SNN (large) &95.60  &3.60  &2863 & 5.01 \\
       LitE-SNN (medium) &94.52  &1.23  &913 & 4.99 \\
	LitE-SNN (small) &91.98  &0.55  &298 & 4.48\\
			\bottomrule
		\end{tabular}}
		\caption{Performance comparison of our joint optimization with sequential optimization.}
		\label{joint} 
	\end{center}
 \end{table}
In this section, we conduct a comparative analysis between our LitE-SNNs and the SNNs optimized using the sequential scheme on CIFAR-10 dataset to validate the effectiveness of our joint optimization scheme. 
The comparison models are derived by sequentially performing neural architecture search (A), pruning (P), quantization (Q) and timestep selection (T). The methods used in each step follow the corresponding scales of LitE-SNNs. 
%We also conduct a comparison of the total time required to design the network from scratch (including search and retraining).

As shown in Table \ref{joint}, the three versions of LitE-SNN consistently outperform models that are derived from sequential optimization, achieving higher accuracy with comparable model sizes and Bit-SynOps. Since each step in sequential models applies the same methods as LitE-SNNs, their inferior accuracy and longer design time demonstrate the effectiveness of our joint optimization. It is worth noting that the small-scale sequential model suffers a substantial accuracy loss. This can be attributed to its high sparsity of 90\%, making it sensitive to parameter adjustments. Small changes in parameter values due to subsequent quantization can significantly impact the network. Whereas our joint optimization considers the intertwined effects of various compressive factors, thereby effectively mitigating accuracy degradation. 

\subsection{Energy Consumption}
%In ANNs, the dot-product operation carries out multiply-accumulate (MAC) operation, which combines addition and multiplication. For SNNs, due to the event-based operation and binary activation, they perform multiplication-free inference, except for the first encoding layer. The addition count is equal to $sf \times T \times A$, where $sf$ is the mean sparsity of firing, $T$ is the time step and $A$ is the addition number in ANN. The number of multiplication operations is set to the FLOPs of the first layer and scaled it by the number of timesteps. 
In this section, we estimate the theoretical energy consumption of our LitE-SNN using the common approach in the neuromorphic community \cite{rathi2021diet,li2021differentiable}. 
The energy is measured in 45nm CMOS technology \cite{horowitz20141}, where the addition operation and multiplication operation cost 0.9 pJ and 3.7 pJ energy respectively. These energy values are based on 32-bit floating-point operations. Since our model uses quantization to a smaller bit width, the reported energy values for our model are overestimates compared to those of other models.

Table \ref{energy} presents the energy consumption of our LitE-SNNs and corresponding ANN counterparts with the same network architecture. Our LitE-SNNs with three scales respectively consume over 6$\times$ lower energy compared with ANN. Compared with our baseline model SpikeDHS, LitE-SNNs achieve an energy reduction of more than 0.58mJ. We also compare our results with AutoSNN, which focuses on designing energy-efficient SNNs. Its energy consumption is comparable with our large-scale LitE-SNN, larger than our small and medium-scale LitE-SNNs. %While we focus on computational energy, our model's reduced size and bit-width may also lower memory access energy, although this aspect relates to hardware design and is outside our study's scope.
We acknowledge that energy consumption is not only from computation but also from memory access, a topic beyond the scope of our study as it is more about hardware design. Nevertheless, it's worth noting that our model's smaller size and lower bit-width can contribute to reducing access-related energy consumption.

\begin{table}[!t]
	\setlength\tabcolsep{2pt}  
	\begin{center}
 \resizebox{0.9\columnwidth}{!}{
		\begin{tabular}{l|ccc}
			\toprule
			Model  &\#Addition(M)& \#Multiplication(M) & Energy (mJ) \\
			\midrule
                ANN (large) &970.12  &970.12 &4.46  \\
                ANN (medium) & 413.53  &  413.53 & 1.90 \\
                ANN (small) & 221.63  & 221.63 & 1.02\\
                \midrule
                AutoSNN \cite{na2022autosnn} & 585.60 & 28.31 & 0.63 \\
                SpikeDHS \cite{che2022differentiable} & 1305.65 &23.89&1.26\\
			LitE-SNN (large)   &705.01  & 11.94 &0.68 \\
                LitE-SNN (medium)   & 271.41 & 4.78 &0.26 \\
                LitE-SNN (small)  & 129.74 & 2.39 &0.13  \\
			\bottomrule
		\end{tabular}}
		\caption{Energy cost comparison for a single forward.}
		\label{energy} 
	\end{center}
 \end{table}

\section{Conclusion}
In this paper, we propose the LitE-SNN to integrate spatio-temporal compression into the automatic network design. We present the CompConv that expands the search space to support pruning and mixed-precision quantization. Meanwhile, we propose the compressive timestep search to identify the optimal number of timesteps to process SNNs. We also formulate a joint optimization algorithm to learn these compression strategies and architecture parameters. Experimental results show that our LitE-SNN exhibits competitive accuracy, compact model size, and fewer computation costs, making it an attractive choice for deployment on resource-constrained edge devices and opening new possibilities for the practical implementation of SNNs in real-world applications.

\section*{Acknowledgements}
This work is supported by the Natural Science Foundation of China (No. 61925603, U1909202) and the Key Research and Development Program of Zhejiang Province in China (2020C03004). This work is also supported by IAF, A*STAR, SOITEC, NXP, National University of Singapore under FD-fAbrICS: Joint Lab for FD-SOI Always-on Intelligent \& Connected Systems (Award I2001E0053) and the Internal Project Fund from Shenzhen Research Institute of Big Data under Grant T00120220002.

\bibliographystyle{named}
\bibliography{ijcai24}

\begin{thebibliography}{}

\bibitem[\protect\citeauthoryear{Cai and Vasconcelos}{2020}]{cai2020rethinking}
Zhaowei Cai and Nuno Vasconcelos.
\newblock Rethinking differentiable search for mixed-precision neural networks.
\newblock In {\em Proceedings of the IEEE/CVF Conference on Computer Vision and Pattern Recognition}, pages 2349--2358, 2020.

\bibitem[\protect\citeauthoryear{Cai \bgroup \em et al.\egroup }{2017}]{cai2017deep}
Zhaowei Cai, Xiaodong He, Jian Sun, and Nuno Vasconcelos.
\newblock Deep learning with low precision by half-wave gaussian quantization.
\newblock In {\em Proceedings of the {IEEE} Conference on Computer Vision and Pattern Recognition}, pages 5918--5926, 2017.

\bibitem[\protect\citeauthoryear{Cai \bgroup \em et al.\egroup }{2019}]{cai2019proxylessnas}
Han Cai, Ligeng Zhu, and Song Han.
\newblock Proxylessnas: Direct neural architecture search on target task and hardware.
\newblock In {\em 7th International Conference on Learning Representations}, 2019.

\bibitem[\protect\citeauthoryear{Che \bgroup \em et al.\egroup }{2022}]{che2022differentiable}
Kaiwei Che, Luziwei Leng, Kaixuan Zhang, Jianguo Zhang, Qinghu Meng, Jie Cheng, Qinghai Guo, and Jianxing Liao.
\newblock Differentiable hierarchical and surrogate gradient search for spiking neural networks.
\newblock {\em Advances in Neural Information Processing Systems}, 35:24975--24990, 2022.

\bibitem[\protect\citeauthoryear{Chen \bgroup \em et al.\egroup }{2021}]{chen2021pruning}
Yanqi Chen, Zhaofei Yu, Wei Fang, Tiejun Huang, and Yonghong Tian.
\newblock Pruning of deep spiking neural networks through gradient rewiring.
\newblock In {\em Proceedings of the Thirtieth International Joint Conference on Artificial Intelligence}, pages 1713--1721, 2021.

\bibitem[\protect\citeauthoryear{Chen \bgroup \em et al.\egroup }{2022}]{chen2022state}
Yanqi Chen, Zhaofei Yu, Wei Fang, Zhengyu Ma, Tiejun Huang, and Yonghong Tian.
\newblock State transition of dendritic spines improves learning of sparse spiking neural networks.
\newblock In {\em International Conference on Machine Learning}, pages 3701--3715, 2022.

\bibitem[\protect\citeauthoryear{Davies and others}{2021}]{davies2021taking}
Mike Davies et~al.
\newblock Taking neuromorphic computing to the next level with loihi2.
\newblock {\em Intel Labs’ Loihi}, 2:1--7, 2021.

\bibitem[\protect\citeauthoryear{Davies \bgroup \em et al.\egroup }{2018}]{davies2018loihi}
Mike Davies, Narayan Srinivasa, Tsung-Han Lin, Gautham Chinya, Yongqiang Cao, Sri~Harsha Choday, Georgios Dimou, Prasad Joshi, Nabil Imam, Shweta Jain, et~al.
\newblock Loihi: A neuromorphic manycore processor with on-chip learning.
\newblock {\em IEEE Micro}, 38(1):82--99, 2018.

\bibitem[\protect\citeauthoryear{Deng \bgroup \em et al.\egroup }{2021}]{deng2021comprehensive}
Lei Deng, Yujie Wu, Yifan Hu, Ling Liang, Guoqi Li, Xing Hu, Yufei Ding, Peng Li, and Yuan Xie.
\newblock Comprehensive snn compression using admm optimization and activity regularization.
\newblock {\em IEEE Transactions on Neural Networks and Learning Systems}, 2021.

\bibitem[\protect\citeauthoryear{Farabet \bgroup \em et al.\egroup }{2012}]{farabet2012comparison}
Cl{\'e}ment Farabet, Rafael Paz, Jose P{\'e}rez-Carrasco, Carlos Zamarre{\~n}o-Ramos, Alejandro Linares-Barranco, Yann LeCun, Eugenio Culurciello, Teresa Serrano-Gotarredona, and Bernabe Linares-Barranco.
\newblock Comparison between frame-constrained fix-pixel-value and frame-free spiking-dynamic-pixel convnets for visual processing.
\newblock {\em Frontiers in Neuroscience}, 6:32, 2012.

\bibitem[\protect\citeauthoryear{Horowitz}{2014}]{horowitz20141}
Mark Horowitz.
\newblock 1.1 computing's energy problem (and what we can do about it).
\newblock In {\em 2014 IEEE international solid-state circuits conference digest of technical papers (ISSCC)}, pages 10--14. IEEE, 2014.

\bibitem[\protect\citeauthoryear{Hu \bgroup \em et al.\egroup }{2021}]{hu2021spiking}
Yangfan Hu, Huajin Tang, and Gang Pan.
\newblock Spiking deep residual networks.
\newblock {\em IEEE Transactions on Neural Networks and Learning Systems}, 34(8):5200--5205, 2021.

\bibitem[\protect\citeauthoryear{Hu \bgroup \em et al.\egroup }{2023}]{hu2023fast}
Yangfan Hu, Qian Zheng, Xudong Jiang, and Gang Pan.
\newblock Fast-snn: Fast spiking neural network by converting quantized ann.
\newblock {\em IEEE Transactions on Pattern Analysis and Machine Intelligence}, 2023.

\bibitem[\protect\citeauthoryear{Kim \bgroup \em et al.\egroup }{2022a}]{kim2022neural}
Youngeun Kim, Yuhang Li, Hyoungseob Park, Yeshwanth Venkatesha, and Priyadarshini Panda.
\newblock Neural architecture search for spiking neural networks.
\newblock In {\em European Conference on Computer Vision}, pages 36--56. Springer, 2022.

\bibitem[\protect\citeauthoryear{Kim \bgroup \em et al.\egroup }{2022b}]{kim2022exploring}
Youngeun Kim, Yuhang Li, Hyoungseob Park, Yeshwanth Venkatesha, Ruokai Yin, and Priyadarshini Panda.
\newblock Exploring lottery ticket hypothesis in spiking neural networks.
\newblock In {\em European Conference on Computer Vision}, pages 102--120. Springer, 2022.

\bibitem[\protect\citeauthoryear{Krizhevsky \bgroup \em et al.\egroup }{2009}]{krizhevsky2009learning}
Alex Krizhevsky, Geoffrey Hinton, et~al.
\newblock Learning multiple layers of features from tiny images.
\newblock 2009.

\bibitem[\protect\citeauthoryear{Li \bgroup \em et al.\egroup }{2021}]{li2021differentiable}
Yuhang Li, Yufei Guo, Shanghang Zhang, Shikuang Deng, Yongqing Hai, and Shi Gu.
\newblock Differentiable spike: Rethinking gradient-descent for training spiking neural networks.
\newblock In {\em Advances in Neural Information Processing Systems}, pages 23426--23439, 2021.

\bibitem[\protect\citeauthoryear{Li \bgroup \em et al.\egroup }{2023}]{li2023input}
Yuhang Li, Abhishek Moitra, Tamar Geller, and Priyadarshini Panda.
\newblock Input-aware dynamic timestep spiking neural networks for efficient in-memory computing.
\newblock In {\em 2023 60th ACM/IEEE Design Automation Conference}, pages 1--6. IEEE, 2023.

\bibitem[\protect\citeauthoryear{Liu \bgroup \em et al.\egroup }{2019}]{liu2019darts}
Hanxiao Liu, Karen Simonyan, and Yiming Yang.
\newblock {DARTS:} differentiable architecture search.
\newblock In {\em International Conference on Learning Representations}, 2019.

\bibitem[\protect\citeauthoryear{Liu \bgroup \em et al.\egroup }{2020}]{liu2020unsupervised}
Qianhui Liu, Gang Pan, Haibo Ruan, Dong Xing, Qi~Xu, and Huajin Tang.
\newblock Unsupervised aer object recognition based on multiscale spatio-temporal features and spiking neurons.
\newblock {\em IEEE Transactions on Neural Networks and Learning Systems}, 31(12):5300--5311, 2020.

\bibitem[\protect\citeauthoryear{Lui and Neftci}{2021}]{lui2021hessian}
Hin~Wai Lui and Emre Neftci.
\newblock Hessian aware quantization of spiking neural networks.
\newblock In {\em International Conference on Neuromorphic Systems 2021}, pages 1--5, 2021.

\bibitem[\protect\citeauthoryear{Ma \bgroup \em et al.\egroup }{2024}]{ma2024darwin3}
De~Ma, Xiaofei Jin, Shichun Sun, Yitao Li, Xundong Wu, Youneng Hu, Fangchao Yang, Huajin Tang, Xiaolei Zhu, Peng Lin, et~al.
\newblock Darwin3: A large-scale neuromorphic chip with a novel isa and on-chip learning.
\newblock {\em National Science Review}, page nwae102, 2024.

\bibitem[\protect\citeauthoryear{Merolla \bgroup \em et al.\egroup }{2014}]{merolla2014million}
Paul~A Merolla, John~V Arthur, Rodrigo Alvarez-Icaza, Andrew~S Cassidy, Jun Sawada, Filipp Akopyan, et~al.
\newblock A million spiking-neuron integrated circuit with a scalable communication network and interface.
\newblock {\em Science}, 345(6197):668--673, 2014.

\bibitem[\protect\citeauthoryear{Na \bgroup \em et al.\egroup }{2022}]{na2022autosnn}
Byunggook Na, Jisoo Mok, Seongsik Park, Dongjin Lee, Hyeokjun Choe, and Sungroh Yoon.
\newblock Autosnn: Towards energy-efficient spiking neural networks.
\newblock In {\em International Conference on Machine Learning}, pages 16253--16269. PMLR, 2022.

\bibitem[\protect\citeauthoryear{Orchard \bgroup \em et al.\egroup }{2021}]{orchard2021efficient}
Garrick Orchard, E~Paxon Frady, Daniel Ben~Dayan Rubin, Sophia Sanborn, Sumit~Bam Shrestha, Friedrich~T Sommer, and Mike Davies.
\newblock Efficient neuromorphic signal processing with loihi 2.
\newblock In {\em 2021 IEEE Workshop on Signal Processing Systems}, pages 254--259. IEEE, 2021.

\bibitem[\protect\citeauthoryear{Rathi and Roy}{2021}]{rathi2021diet}
Nitin Rathi and Kaushik Roy.
\newblock Diet-snn: A low-latency spiking neural network with direct input encoding and leakage and threshold optimization.
\newblock {\em IEEE Transactions on Neural Networks and Learning Systems}, 2021.

\bibitem[\protect\citeauthoryear{Rathi \bgroup \em et al.\egroup }{2018}]{rathi2018stdp}
Nitin Rathi, Priyadarshini Panda, and Kaushik Roy.
\newblock Stdp-based pruning of connections and weight quantization in spiking neural networks for energy-efficient recognition.
\newblock {\em IEEE Transactions on Computer-Aided Design of Integrated Circuits and Systems}, 38(4):668--677, 2018.

\bibitem[\protect\citeauthoryear{Rueckauer \bgroup \em et al.\egroup }{2017}]{rueckauer2017conversion}
Bodo Rueckauer, Iulia-Alexandra Lungu, Yuhuang Hu, Michael Pfeiffer, and Shih-Chii Liu.
\newblock Conversion of continuous-valued deep networks to efficient event-driven networks for image classification.
\newblock {\em Frontiers in Neuroscience}, 11:682, 2017.

\bibitem[\protect\citeauthoryear{Sehwag \bgroup \em et al.\egroup }{2020}]{sehwag2020hydra}
Vikash Sehwag, Shiqi Wang, Prateek Mittal, and Suman Jana.
\newblock Hydra: Pruning adversarially robust neural networks.
\newblock {\em Advances in Neural Information Processing Systems}, 33:19655--19666, 2020.

\bibitem[\protect\citeauthoryear{Shen \bgroup \em et al.\egroup }{2023}]{shen2023esl}
Jiangrong Shen, Qi~Xu, Jian~K Liu, Yueming Wang, Gang Pan, and Huajin Tang.
\newblock Esl-snns: An evolutionary structure learning strategy for spiking neural networks.
\newblock In {\em Proceedings of the AAAI Conference on Artificial Intelligence}, volume~37, pages 86--93, 2023.

\bibitem[\protect\citeauthoryear{Srinivasan and Roy}{2019}]{srinivasan2019restocnet}
Gopalakrishnan Srinivasan and Kaushik Roy.
\newblock Restocnet: Residual stochastic binary convolutional spiking neural network for memory-efficient neuromorphic computing.
\newblock {\em Frontiers in Neuroscience}, 13:189, 2019.

\bibitem[\protect\citeauthoryear{Wang \bgroup \em et al.\egroup }{2020}]{wang2020apq}
Tianzhe Wang, Kuan Wang, Han Cai, Ji~Lin, Zhijian Liu, Hanrui Wang, Yujun Lin, and Song Han.
\newblock Apq: Joint search for network architecture, pruning and quantization policy.
\newblock In {\em Proceedings of the IEEE/CVF Conference on Computer Vision and Pattern Recognition}, pages 2078--2087, 2020.

\bibitem[\protect\citeauthoryear{Wang \bgroup \em et al.\egroup }{2023}]{wang2023event}
Yang Wang, Bo~Dong, Yuji Zhang, Yunduo Zhou, Haiyang Mei, Ziqi Wei, and Xin Yang.
\newblock Event-enhanced multi-modal spiking neural network for dynamic obstacle avoidance.
\newblock In {\em Proceedings of the 31st ACM International Conference on Multimedia}, pages 3138--3148, 2023.

\bibitem[\protect\citeauthoryear{Warden}{2018}]{warden2018speech}
Pete Warden.
\newblock Speech commands: A dataset for limited-vocabulary speech recognition.
\newblock {\em arXiv preprint arXiv:1804.03209}, 2018.

\bibitem[\protect\citeauthoryear{Wei \bgroup \em et al.\egroup }{2024}]{wei2024event}
Wenjie Wei, Malu Zhang, Jilin Zhang, Ammar Belatreche, Jibin Wu, Zijing Xu, Xuerui Qiu, Hong Chen, Yang Yang, and Haizhou Li.
\newblock Event-driven learning for spiking neural networks.
\newblock {\em arXiv preprint arXiv:2403.00270}, 2024.

\bibitem[\protect\citeauthoryear{Wu \bgroup \em et al.\egroup }{2021}]{wu2021tandem}
Jibin Wu, Yansong Chua, Malu Zhang, Guoqi Li, Haizhou Li, and Kay~Chen Tan.
\newblock A tandem learning rule for effective training and rapid inference of deep spiking neural networks.
\newblock {\em IEEE Transactions on Neural Networks and Learning Systems}, 2021.

\bibitem[\protect\citeauthoryear{Xu \bgroup \em et al.\egroup }{2023}]{xu2023constructing}
Qi~Xu, Yaxin Li, Jiangrong Shen, Jian~K Liu, Huajin Tang, and Gang Pan.
\newblock Constructing deep spiking neural networks from artificial neural networks with knowledge distillation.
\newblock In {\em Proceedings of the IEEE/CVF Conference on Computer Vision and Pattern Recognition}, pages 7886--7895, 2023.

\bibitem[\protect\citeauthoryear{Yan \bgroup \em et al.\egroup }{2024}]{yan2024efficient}
Jiaqi Yan, Qianhui Liu, Malu Zhang, Lang Feng, De~Ma, Haizhou Li, and Gang Pan.
\newblock Efficient spiking neural network design via neural architecture search.
\newblock {\em Neural Networks}, page 106172, 2024.

\bibitem[\protect\citeauthoryear{Yang \bgroup \em et al.\egroup }{2022}]{yang2022deep}
Qu~Yang, Qi~Liu, and Haizhou Li.
\newblock Deep residual spiking neural network for keyword spotting in low-resource settings.
\newblock In {\em INTERSPEECH}, pages 3023--3027, 2022.

\bibitem[\protect\citeauthoryear{Y{\i}lmaz \bgroup \em et al.\egroup }{2020}]{yilmaz2020deep}
Emre Y{\i}lmaz, Ozg{\"u}r~Bora Gevrek, Jibin Wu, Yuxiang Chen, Xuanbo Meng, and Haizhou Li.
\newblock Deep convolutional spiking neural networks for keyword spotting.
\newblock In {\em INTERSPEECH}, pages 2557--2561, 2020.

\bibitem[\protect\citeauthoryear{Zhang \bgroup \em et al.\egroup }{2016}]{zhang2016cambricon}
Shijin Zhang, Zidong Du, Lei Zhang, Huiying Lan, Shaoli Liu, Ling Li, Qi~Guo, Tianshi Chen, and Yunji Chen.
\newblock Cambricon-x: An accelerator for sparse neural networks.
\newblock In {\em 2016 49th Annual IEEE/ACM International Symposium on Microarchitecture}, pages 1--12. IEEE, 2016.

\bibitem[\protect\citeauthoryear{Zhang \bgroup \em et al.\egroup }{2021}]{zhang2021rectified}
Malu Zhang, Jiadong Wang, Jibin Wu, Ammar Belatreche, Burin Amornpaisannon, Zhixuan Zhang, et~al.
\newblock Rectified linear postsynaptic potential function for backpropagation in deep spiking neural networks.
\newblock {\em IEEE Transactions on Neural Networks and Learning Systems}, 33(5):1947--1958, 2021.

\bibitem[\protect\citeauthoryear{Zheng \bgroup \em et al.\egroup }{2021}]{zheng2021going}
Hanle Zheng, Yujie Wu, Lei Deng, Yifan Hu, and Guoqi Li.
\newblock Going deeper with directly-trained larger spiking neural networks.
\newblock In {\em Proceedings of the AAAI Conference on Artificial Intelligence}, volume~35, pages 11062--11070, 2021.

\end{thebibliography}

\end{document}